\documentclass[lettersize,journal]{IEEEtran}

\usepackage{amsmath,amsfonts}
\usepackage{algorithm}
\usepackage{array}
\usepackage[caption=false,font=normalsize,labelfont=sf,textfont=sf]{subfig}
\usepackage{textcomp}
\usepackage{stfloats}
\usepackage{url}
\usepackage{verbatim}
\usepackage{graphicx}
\usepackage{cite}

\usepackage{subfig}
\usepackage{multirow}
\usepackage{amssymb}
\usepackage{makecell}
\usepackage{xcolor}

\usepackage{algpseudocode}
\usepackage{multirow}

\usepackage{mathtools} % zx
\usepackage{booktabs} % zx
\usepackage{amsmath} % zx
\usepackage{amsthm} % zx
\usepackage{color} % zx
\usepackage{colortbl} % zx
\usepackage{bm} % zx
\usepackage{float} 

\definecolor{rblue}{rgb}{0,0.5,1}
\usepackage[bookmarks=false]{hyperref}
\usepackage{cleveref}
\hypersetup{colorlinks=true,linkcolor=red,citecolor=rblue}

\begin{document}

\title{Cross-modal Context-aware Learning for Visual Prompt Guided Multimodal Image Understanding in Remote Sensing}

\author{
Xu~Zhang,
Jiabin Fang,
Zhuoming Ding,
Jin~Yuan\IEEEauthorrefmark{1},
Xuan Liu,
Qianjun Zhang\IEEEauthorrefmark{1},
and Zhiyong~Li
\thanks{X. Zhang, J. Fang, Z. Ding, J. Yuan, and X. Liu are with the College of Computer Science and Electronic Engineering, Hunan University, Changsha 410082, China.}
\thanks{Z. Li is with the School of Robotics and the National Engineering Research Center of Robot Visual Perception and Control Technology, Hunan University, Changsha 410082, China.}
\thanks{Q. Zhang is with the School of Computing and Artificial Intelligence, Southwest Jiaotong University, Sichuan 611756, China.}
\thanks{\IEEEauthorrefmark{1}Corresponding authors: Jin Yuan and Qianjun Zhang. (E-mail: yuanjin@hnu.edu.cn, zqjblue@foxmail.com.)}
}

% The paper headers
% \markboth{Journal of \LaTeX\ Class Files,~Vol.~14, No.~8, August~2021}%
% {Shell \MakeLowercase{\textit{et al.}}: A Sample Article Using IEEEtran.cls for IEEE Journals}

% \IEEEpubid{0000--0000/00\$00.00~\copyright~2021 IEEE}
% % Remember, if you use this you must call \IEEEpubidadjcol in the second
% % column for its text to clear the IEEEpubid mark.

\maketitle

\begin{abstract}
Recent advances in image understanding have enabled methods that leverage large language models for multimodal reasoning in remote sensing. However, existing approaches still struggle to steer models to the user-relevant regions when only simple, generic text prompts are available. Moreover, in large-scale aerial imagery many objects exhibit highly similar visual appearances and carry rich inter-object relationships, which further complicates accurate recognition.
To address these challenges, we propose Cross-modal Context-aware Learning for Visual Prompt–Guided Multimodal Image Understanding (CLV-Net). CLV-Net lets users supply a simple visual cue—a bounding box—to indicate a region of interest, and uses that cue to guide the model to generate correlated segmentation masks and captions that faithfully reflect user intent. Central to our design is a Context-Aware Mask Decoder that models and integrates inter-object relationships to strengthen target representations and improve mask quality. In addition, we introduce a Semantic and Relationship Alignment module: a Cross-modal Semantic Consistency Loss enhances fine-grained discrimination among visually similar targets, while a Relationship Consistency Loss enforces alignment between textual relations and visual interactions.
Comprehensive experiments on two benchmark datasets show that CLV-Net outperforms existing methods and establishes new state-of-the-art results. The model effectively captures user intent and produces precise, intention-aligned multimodal outputs.

\end{abstract}


\section{Introduction}
\label{sec:intro}

\begin{figure*}[!t]
	\centering
	\includegraphics[width=1\linewidth]{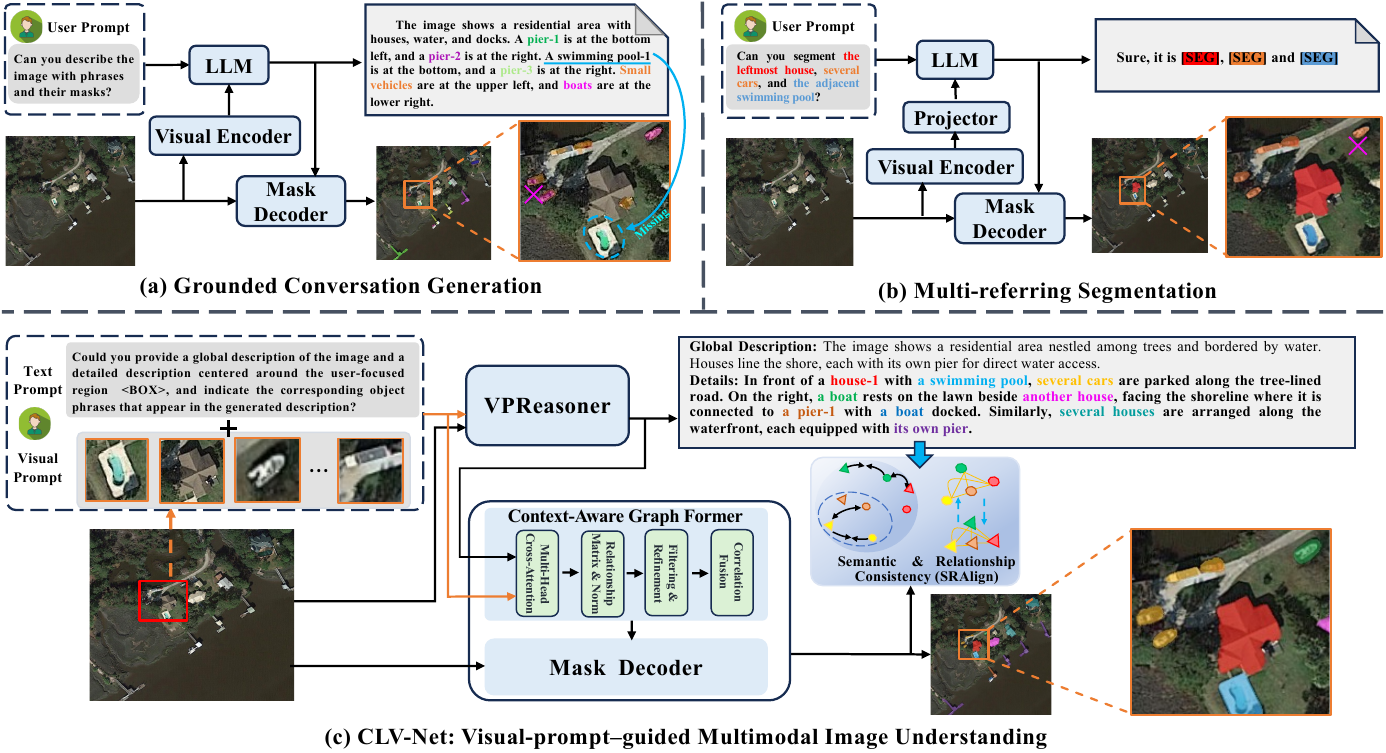}
	\caption{Framework Comparison between our method and the existing methods. 
    Grounded Conversation Generation (subfigure a) directly generates a global summary on an image, while Multi-referring Segmentation (subfigure a) allows detail VQA and segmentation on local regions conditioned on a textual prompt. Both approaches predict wrong labels for masks due to the similar visual appearances between objects. Differently, our approach  (subfigure c) receives a simple visual prompt embedded into a textual prompt to generate a concise global and detailed local descriptions along with accurate masks corresponding to caption tokens. Underlined text denotes predicted words with missing masks, while crosses indicate mismatches between the generated text and the target.}
    \label{fig:motivation}
\end{figure*}

\IEEEPARstart{I}{mage} understanding — the extraction of essential semantic information from images via visual and textual representations — underpins a wide range of visual tasks such as object detection~\cite{dai2021dynamic}, image captioning~\cite{li2022diffusion,luo2023semantic}, and segmentation~\cite{yu2020sprnet, seem_zou2023segment,ren2025interactive}. These technologies have found broad application in human–computer interaction~\cite{zhang2024pvpuformer, SAM_kirillov2023segment}, autonomous driving~\cite{li2023bi, lin2024echotrack, wei2024editable}, and medical diagnosis~\cite{liu2025generalist}.
Recent advances in large language models (LLMs), and especially multimodal LLMs, have shifted image understanding from purely unimodal perception toward multimodal, collaborative comprehension. By exploiting cross-modal correlations and the strong reasoning capabilities of MLLMs, these approaches not only improve accuracy but also reveal deeper semantic structures in scenes, enabling richer, context-aware interpretations of visual content and delivering notable gains in natural scene understanding~\cite{he2024ma, wang2024semantic, lai2024lisa, zhang2024groundhog, rasheed2023glamm, zhang2025sgdiff}.

Unlike natural images, remote sensing imagery covers vast areas and often contains targets with subtle, low-contrast features, which greatly increases the difficulty of precisely focusing on and interpreting scene content. Recent work has begun applying multimodal large language models (MLLMs) to collaborative understanding in remote sensing, including GeoPix~\cite{ou2025geopix}, GeoChat~\cite{kuckreja2024geochat}, GeoGround~\cite{zhou2024geoground}, and GeoPixel~\cite{shabbir2025geopixel}.
For example, GeoPixel~\cite{shabbir2025geopixel} uses textual prompts to guide an LLM in producing global semantic descriptions, which are then leveraged for reasoning-aware segmentation to generate both textual summaries and corresponding target masks (see ~\Cref{fig:motivation}(a)). Although this yields useful high-level scene semantics, it lacks fine-grained understanding of local regions. To address that gap, GeoPix~\cite{ou2025geopix} augments textual descriptions to support multi-referring segmentation and visual question answering, enabling more detailed localization of local areas (see ~\Cref{fig:motivation}(b)). Nevertheless, GeoPix relies on experts to scan vast imagery, identify regions of interest, and craft referring expressions, which imposes substantial cognitive and interaction burdens on users.

This paper presents a visual-prompt–guided multimodal framework for remote-sensing image understanding. In addition to a global textual prompt, our system accepts a simple visual prompt—e.g., a bounding box—that specifies a region of interest. The model then generates a compact global semantic caption plus several detailed local descriptions, each paired with a segmentation mask (see~\Cref{fig:motivation}(c)). By combining visual and textual cues, the approach enables precise, user-focused interpretation while substantially reducing user effort. It also strikes a balance between comprehensive global understanding and fine-grained local analysis.

To implement this framework, we propose Cross-modal Context-aware Learning for Visual-prompt–guided Multimodal Image Understanding in Remote Sensing (CLV-Net), which comprises two principal modules: a Visual-Prompt Scene Reasoner (VPReasoner) and a Context-Aware Mask Decoder (CMDecoder). As shown in ~\Cref{fig:motivation}(c), the VPReasoner ingests a global image together with visual and textual prompts, fuses these bimodal cues, and conditions an LLM to generate hierarchical captions — a concise global summary alongside detailed local descriptions. Building on this, the CMDecoder explicitly models cross-modal relationships between textual and visual representations and captures contextual dependencies among objects. Cross-modal modeling yields precise alignment between mask regions and caption tokens, while contextual modeling reduces object classification errors by exploiting neighboring object context. Consequently, the CMDecoder produces mask outputs whose object labels accurately correspond to the generated captions. To narrow the representation gap between caption tokens and object masks, we introduce two novel training objectives: a Semantic Consistency loss and a Relationship Consistency loss. The Semantic Consistency loss pulls cross-modal representations of the same object closer while pushing apart representations of different objects. The Relationship Consistency loss enforces cross-modal relational agreement by leveraging semantic relationships in the textual domain to alleviate category ambiguity during visual decoding. Together, these losses improve the model’s ability to discriminate between objects and ensure consistent correspondence between caption tokens and masks. Extensive evaluations on the remote-sensing dataset GeoPixelD~\cite{shabbir2025geopixel} and the natural-image dataset GranD~\cite{rasheed2023glamm} demonstrate the effectiveness of our framework, which achieves state-of-the-art performance compared with existing methods.

In summary, the main contributions of this work are as follows:
\begin{itemize}
\item We propose Cross-modal Context-aware Learning for Visual-prompt–guided Multimodal Image Understanding in Remote Sensing (CLV-Net). Given a simple visual prompt that specifies a local region in the image, CLV-Net enables generation of a concise global summary together with detailed local descriptions and their corresponding masks—reducing user effort while producing richer, intention-aligned semantic outputs.
\item We design a Context-Aware Mask Decoder that explicitly models and fuses inter-object semantic relationships to enhance object representations and improve mask quality. To further bridge modalities, we introduce two novel training objectives that enforce cross-modal representation and relationship consistency, yielding precise correspondence between caption tokens and masks.
\item Extensive experiments on the GeoPixelD remote-sensing dataset and the GranD natural-image dataset demonstrate the effectiveness and generalizability of our approach, achieving new state-of-the-art results across diverse domains.
\end{itemize}

\section{Related work}
\label{sec:related_work}

\subsection{Multimodal Large Language Models}
Large language models (LLMs) have emerged as powerful tools for bridging vision and language, enabling joint reasoning across modalities. Early efforts extended language models with visual backbones to perform tasks such as captioning~\cite{devlin2019bert, zhang2024earthgpt}, visual question answering~\cite{zheng2023judging, liu2023visual, zhang2023multistep}, and dialogue-based scene understanding~\cite{li2022diffusion, he2024ma}. 
More recent research has evolved from single-modality processing toward Multimodal Large Language Models (MLLMs)~\cite{li2023blip,zhu2023minigpt,wang2023visionllm}, which integrate visual and linguistic understanding within a unified framework to achieve fine-grained grounding and pixel-level reasoning through the reasoning capabilities of LLMs. 
For instance, GSVa~\cite{xia2024gsva} introduces generalized segmentation via multimodal LLMs, and Groma~\cite{ma2024groma} proposes localized visual tokenization for fine-grained grounding. 
LISA~\cite{lai2024lisa} is the first framework to leverage semantic text generated by LLMs for reasoning-based segmentation, inaugurating the LLM-based collaborative understanding paradigm. Building upon LISA, GLaMM~\cite{rasheed2023glamm} extends this paradigm to enable multi-object collaborative understanding scenes, producing coordinated outputs across multiple targets within an image.
PixelLM and GROUNDHOG~\cite{ren2024pixellm, zhang2024groundhog} further advance pixel-level reasoning with LLMs, while STVGBert~\cite{su2021stvgbert} extends visual grounding into spatio-temporal video contexts.
More recently, OMG-LLaVA~\cite{zhang2024omg} unifies reasoning across image-level, object-level, and pixel-level granularity, demonstrating the growing versatility of MLLMs in multimodal understanding.

Despite these advances, most existing methods rely on generic textual prompts for grounding, which often lack the granularity and precision needed to capture user intent in complex visual scenes. 
In contrast to region-based image understanding methods~\cite{wang2023caption,huang2024segment}, which restrict the output strictly to the content within a bounding box, our method leverages the bounding box as an anchor to locate the region of interest while generating relation-aware descriptions that extend beyond the box. This enables the model not only to focus on the targets of interest but also to capture connections between these targets and the surrounding context.

\subsection{Remote Sensing Multimodal LLMs}
Remote sensing (RS) presents unique challenges for multimodal reasoning due to its large-scale, high-resolution imagery and dense object distributions. The substantial domain gap between natural and RS images limits the direct applicability of existing LLM-based methods, often leading to degraded performance. To bridge this gap, recent efforts have extended multimodal LLMs to the RS domain and developed specialized datasets.

RemoteCLIP~\cite{liu2024remoteclip} pioneers a vision-language foundation model for RS, learning semantically rich and transferable representations through large-scale pre-training on diverse RS datasets. 
LRS-VQA~\cite{luo2025large} introduces text-guided token pruning with a Dynamic Image Pyramid for efficient processing of large RS imagery and establishes a benchmark for RS question answering. 
EarthGPT~\cite{zhang2024earthgpt} integrates multisensor data, including optical, SAR, and infrared, through visual-enhanced perception and mutual comprehension mechanisms, supported by the MMRS-1M dataset for instruction-based learning.
RSGPT~\cite{hu2025rsgpt} emphasizes large-scale data curation and constructs the RSICap dataset to enhance vision-language alignment for RS captioning. 
Furthermore, SkyEyeGPT~\cite{zhan2025skyeyegpt} unifies multiple RS vision-language tasks with instruction-tuning across single- and multi-task conversations to improve adaptability. 
SkySenseGPT~\cite{luo2024skysensegpt} advances fine-grained reasoning through the FIT-RS dataset, achieving state-of-the-art performance across RS benchmarks. SkySense V2~\cite{zhang2025skysense} further refines this paradigm with a unified Transformer backbone and self-supervised pre-training, leveraging adaptive patch merging and learnable modality prompts for efficient multimodal learning.
Graph-based models~\cite{tang2025remote,li2025star} such as STAR enhance contextual reasoning by modeling spatial and semantic relations among RS objects. 
Large-scale datasets like RS5M and GeoRSCLIP~\cite{zhang2024rs5m} further support general-purpose RS multimodal training and evaluation. Collectively, these developments establish a comprehensive foundation for vision-language understanding in remote sensing and drive the evolution of remote sensing multimodal LLMs.

\subsection{Visual Grounding LLMs in Remote Sensing}
Relying solely on textual descriptions from large language models (LLMs) fails to provide users with rich and detailed explanations of image content. To enhance users' understanding of remote sensing (RS) imagery, recent works have extended RS LLMs to include pixel-level visual grounding and segmentation, providing text descriptions alongside corresponding target regions for improved accuracy.
GeoChat~\cite{kuckreja2024geochat} is the first versatile remote sensing vision-language model (VLM) offering multitask conversational capabilities with high-resolution RS images. It can answer image-level queries and engage in region-specific dialogues, grounding objects by referencing their spatial coordinates. 
GeoGround~\cite{zhou2024geoground} introduces a unified framework for HBB, OBB, and mask generation tasks, using a Text-Mask technique for pixel-level output. 
GeoPix~\cite{ou2025geopix} extends image-level tasks, such as captioning and visual question answering, to pixel-level segmentation, facilitating multi-referring segmentation tasks. However, it requires users to provide target descriptions, increasing the interaction burden. 
GeoPixel~\cite{shabbir2025geopixel} is the first end-to-end high-resolution RS LLM supporting pixel-level grounding, generating interleaved masks for fine-grained visual perception. 
Despite these advances, they largely rely on generic prompt templates, which are insufficient for capturing user-specific requirements in complex, object-dense scenes. Moreover, most existing frameworks overlook fine-grained inter-object relationships and struggle to disambiguate visually similar targets, which are particularly prevalent in RS imagery. Distinct from prior approaches, our framework incorporates compact visual cues (bounding boxes) to guide model attention toward user-selected regions. Furthermore, by explicitly modeling inter-object relationships and introducing semantic–relational alignment, our method bridges the gap between user intent and multimodal outputs, enabling precise, controllable segmentation and captioning in complex RS scenarios.

% \section{PROPOSED FRAMEWORK: SRC-Net}
\section{Method}
\label{sec:formatting}
We first introduce an overview of the CLV-Net framework in Sec.~\ref{sec:method_Overview}, outlining its components and their interactions. In Sec. \ref{sec:method_VPReasoner} and \ref{sec:method_CMDecoder}, we detail the VPReasoner and CMDecoder modules, which are designed to generate region-specific captions and refine cross-modal correlations for precise mask generation. Finally, Sec. \ref{sec:method_SRAlign} focuses on the SRAlign module, discussing its role in enforcing fine-grained semantic contrastive learning and ensuring relationship consistency between visual and textual representations. 

\subsection{Overview}
\label{sec:method_Overview}
\begin{figure*}[ht]
	\centering
	\includegraphics[width=1\linewidth]{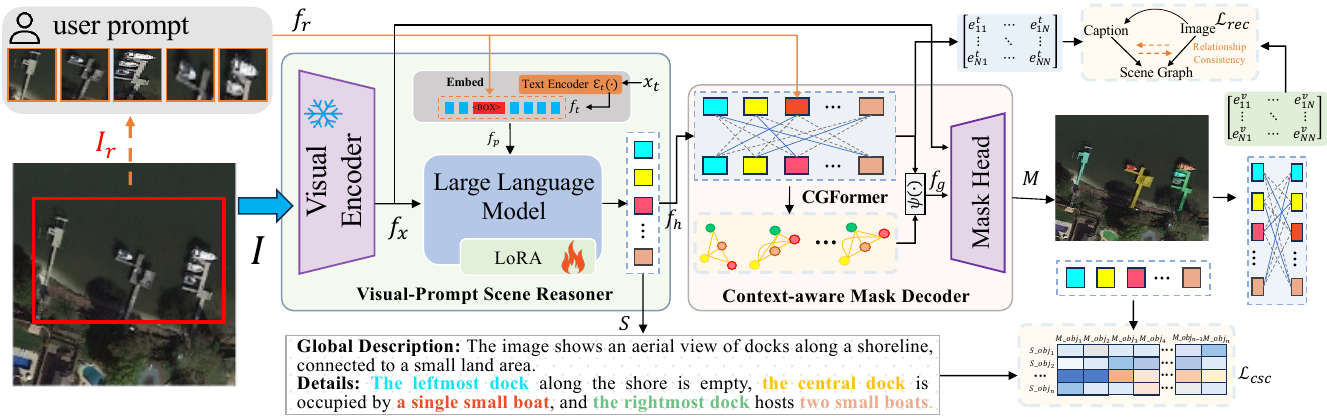}
	\caption{Overview of the proposed Cross-modal Context-aware Learning for Visual Prompt Guided Multimodal Image Understanding in Remote Sensing (CLV-Net), comprising three components: a Visual-Prompt Scene Reasoner (VPReasoner) that uses a bounding box to guide content generation for the user-specified region; a Context-Aware Mask Decoder (CMDecoder) that incorporates inter-object relationships to enhance target representations and mask quality; and a Semantic and Relationship Alignment Module (SRAlign) that enforces fine-grained discrimination via Cross-modal Semantic Consistency Loss and maintains visual-textual relationship consistency via Relationship Consistency Loss.}
    \label{fig:networks}
\end{figure*}

Given  an image $\boldsymbol{I} \in \mathbb{R}^{H \times W \times 3}$ with height $H$ and width $W$, along with a user-provided prompt $P_o$, such as a box bounding a region of interest containing a target set $O$, CLV-Net aims to generate a corresponding pair of caption $S$ and segmentation masks $M$ that precisely aligns with the user’s intention. Here, $S$ describes the semantic information of the targets in $O$, and $M$ represents the corresponding masks for them. To achieve this goal, we design a novel "\textbf{C}ross-modal Context-aware \textbf{L}earning for \textbf{V}isual Prompt Guided Multimodal Image Understanding in Remote Sensing (CLV-Net), which comprises three main components: a Visual-Prompt Scene Reasoner (VPReasoner), a Context-Aware Mask Decoder (CMDecoder) and a Semantic and Relationship Alignment (SRAlign) module. As illustrated in \Cref{fig:networks}, our approach first employs the VPReasoner to generate a caption on the condition of the user-specified region. Next, the CMDecoder deeply explores cross-modal correlations between objects, yielding precise mask results.
Finally, the SRAlign module enhances fine-grained discrimination between visually similar targets through the Cross-modal Semantic Consistency Loss, while maintaining visual-textual relationship consistency via the Relationship Consistency Loss.
% Unlike prior methods that rely on generic prompts and fail to capture user-specific needs, our approach uses a simple visual cue to accurately capture user intent, model inter-object relationships, and resolve fine-grained visual distinctions, enabling the generation of targeted bimodal outputs that closely align with user requirements while minimizing user effort.

%-------------------------------------------------------------------------
\subsection{Visual-Prompt Scene Reasoner}
\label{sec:method_VPReasoner}
Given a global image $\boldsymbol{I}$ and a bounding box $B$ specifying the region of interest $I_r$ by a user, our VPReasoner generates a textual description $S$ that faithfully meet the user's intention. 
Specifically, we employ ViT-H/14 CLIP~\cite{radford2021learning} as the visual encoder $\mathcal{E}_{v}$ to extract global visual features $f_x$ from $\boldsymbol{I}$.
Furthermore, we utilize a pretrained object detector $\mathcal{E}_{d}$ ~\cite{anderson2018bottom, ren2015faster} to extract $K$ object features $f_r \in \mathbb{R}^{K \times D} =\{O_{i}\}_{l=1}^K$ from $I_r$, along with their corresponding spatial coordinates $\{b_{i}\}_{l=1}^K$.
These visual features are then mapped into the language space through a two-layer MLP, serving as a vision-to-language projection module ${\phi}_{l}$:
\begin{equation}
{f}_{x} = {\phi}_{l}(\mathcal{E}_{v}(I)),\, {f}_{r} = {\phi}_{l}(\mathcal{E}_{d}(I_{r})).
\label{eq:1}
\end{equation}
Meanwhile, we formulate a text query $x_t$ as ``Could you provide a global description of the image and a detailed description centered around the user-focused region  \verb|<BOX>|, and indicate the corresponding object phrases that appear in the generated description?''. This query is then passed to a text encoder $\mathcal{E}_{t}$ to obtain discrete embedding $f_t \in \mathbb{R}^{L \times D}$, where $L$ represents the length of the text tokens. The formula is as follows:
\begin{equation}
f_{t} = \mathcal{E}_{t}(x_{t}).
\label{eq:text_encoder}
\end{equation}
Here, $f_t$ contains natural language tokens and special placeholders \verb|<BOX>|. We embed $f_r$ into \verb|<BOX>|, yielding fused prompt embedding $f_p \in \mathbb{R}^{N \times D}$ that integrates localized visual information as: 
\begin{equation}
f_{p} = \text{Embed}(f_t, f_r).
\label{eq:embed}
\end{equation}
Subsequently, the global visual features $f_x$ and the fused prompt embeddings $f_p$ are fed into a LLM $\mathcal{L}$ to generate a caption $S$, along with the object representation $f_h \in \mathbb{R}^{N \times D}$ containing $N$ objects in $S$ represented as a $D$-dimensional vector for each object:
\begin{equation}
S, f_h = \mathcal{L}( f_x, f_p ).
\label{eq:2}
\end{equation}
In the implementation, we extract $f_h$ from the last transformer layer in the LLM.

Finally, we train the VPReasoner by using a captioning loss $\mathcal{L_{\text{caption}}}$, which is formulated as:
\begin{equation}
\mathcal{L_{\text{caption}}} = \text{CE}(\hat{S}, S),
\label{eq:3}
\end{equation}
where \text{CE} represents the cross-entropy operation and $\hat{S}$ denotes the ground-truth textual labels. To efficiently adapt the LLM to our task, we employ low-rank adaptation (LoRA)~\cite{hu2022lora} for parameter-efficient fine-tuning.

Instead of directly adopting large language models (LLMs) to generate captions on global images, our method enables users to point out a region of interest to generate precise and context-relevant captions.
Technically, we leverage object features as prompts to help the LLM in generating precise, intent-aligned captions, while simultaneously producing object-level textual features to support the following instance mask generation and fine-grained cross-modal alignment.

This visual prompt guided interaction reduces user effort and effectively mitigates the difficulties posed by the wide spatial coverage and high resolution of remote sensing imagery, which otherwise make it hard to specify targets or regions using brief textual instructions.
In contrast to region-based methods~\cite{wang2023caption,huang2024segment} that restrict outputs to the interior of the user box and overlook salient cues from surrounding context, our approach integrates both in-box evidence and global scene information to yield more accurate and context-aware results.

\subsection{Context-Aware Mask Decoder} 
\label{sec:method_CMDecoder}
Objects in remote sensing images often exhibit visual indistinguishability, leading to misalignments between textual referring expressions and visual instances during mask generation. For example, in Fig.~\ref{fig:motivation} (a), two cars located on the upper-left shore are mistakenly identified as boats, likely due to visual interference from the nearby boat on the right. 
To mitigate such errors, we propose a Context-Aware Mask Decoder (CMDecoder) designed to disentangle visually similar objects and enhance cross-modal alignment. 
The CMDecoder consists of a Context-Aware Graph Former (CGFormer) and a mask decoder. 
By explicitly exploring the correlation among objects and integrating these correlations into the feature representation, CMDecoder effectively distinguishes visually similar objects and enhances the correspondence accuracy between textual phrases and visual objects.
% CMDecoder explicitly explores the correlation among objects and integrates this correlation into feature representation, thereby enabling the better distinction of different objects and enhancing the correspondence accuracy between text phrase and visual objects.

Concretely, CGFormer first deeply explores cross-modal associations between the text-level object features $f_h$ and the visual features $f_r$, yielding an enhanced text-level object feature $f_o \in \mathbb{R}^{N \times D}$ that accurately captures the cross-modal properties of the referenced targets from both modalities: 
\begin{equation}
f_{o} = {\phi}_{o}({f}_{h} + \text{MHCA}({f}_{h}, {f}_{r}, {f}_{r})),
\label{eq:mhca1}
\end{equation}
where $\text{MHCA}$ represents the multi-head cross-attention layers, and ${\phi}_{o}$ denotes a linear transformation. Next, our approach constructs a relationship matrix $\mathcal{R_{\text{e}}} \in \mathbb{R}^{N \times N}$ to explicitly model the inter-object dependencies, which is formulated as follows:
\begin{equation}
% \mathcal{R_{\text{e}}} = \text{Matrix}(f_o, {f}^{\top}_o).
\mathcal{R_{\text{e}}} = \text{Norm}({\phi}_{h}(f_o) \cdot {\phi}_{v}({f}_o^{\top})),
\label{eq:4}
\end{equation}
where ${\phi}_{h}$ and ${\phi}_{v}$ represent two linear transformations, $Norm$ is a normalization operation to calculate normalized relation weights on each object. The high value $r_{ij}$ in $\mathcal{R_{\text{e}}}$ indicates the strong correlation between the objects $i$ and $j$. We then fuse this object correlation information into $f_{o}$ to generate a relationship-augmented object feature as: 
\begin{equation}
f_e = {\phi}_{e}((\mathcal{R_{\text{e}}} + \text{I} ) f_o),
\label{eq:5}
\end{equation}
where  $I$ is the identity matrix for skip connections, and ${\phi}_e$ is an MLP operation layer.
% Finally, we integrate $f_e$ into the object node features $f_o$ using the adjacency matrix $A_{o}$ derived from the relationship matrix $\mathcal{R_{\text{e}}}$, yielding the refined object representations $f_g \in \mathbb{R}^{N \times D}$. To construct $A_o$, we employ a thresholding strategy whereby an $A_{o}{(i,j)}$ is set to 1 if the corresponding relation score $r_{i,j}$ in $\mathcal{R_{\text{e}}}$ exceeds a predefined threshold $\tau$, and 0 otherwise. This procedure binarizes the dense correlation structure into a sparse graph representation, thereby retaining only the most salient inter-object dependencies while effectively filtering out weak or noisy associations.
We observe that the relation matrix $\mathcal{R_{\text{e}}}$ contains both meaningful connections and spurious, low-magnitude associations. To suppress weak or noisy relations, we derive a binary object adjacency matrix $A_o$ via a thresholding strategy. 
Specifically, for each pair of objects $(i, j)$, we set $A_o(i, j) = 1$ if the corresponding relation score $e_{i,j}$ in $\mathcal{R_{\text{e}}}$ exceeds a predefined threshold $\tau$; otherwise, $A_o(i, j) = 0$.
This procedure binarizes the dense correlation structure into a sparse graph representation, thereby retaining only the most salient inter-object dependencies while effectively filtering out weak or noisy associations.
Subsequently, we leverage the refined adjacency matrix $A_o$ to integrate the relation-enhanced features $f_e$ into the object node representations $f_o$. This yields the final context-aware object representations $f_g \in \mathbb{R}^{N \times D}$, which encode both visual semantics and robust relational cues.
This process is expressed as follows:
\begin{equation}
f_g = \psi({f}_{o}, {\phi}_{g}(f_e \cdot A_{o})).
\label{eq:6}
\end{equation}
where $\psi(,)$ indicates the element-wise addition of two components followed by an MLP transformation, and ${\phi}_g$ is an MLP operation layer.

% \textbf{Mask Generation.} 
For mask generation, we use the SAM2 decoder~\cite{ravi2024sam} to output masks. The decoder $\mathcal{D}$ takes as input the global visual feature $f_x$, and the context-aware object features $f_{g}$ to produce binary region masks $M \in \mathbb{R}^{N\times H\times W}$. The mask loss $\mathcal{L_{\text{Mask}}}$ combines Dice loss and binary cross-entropy (BCE) loss to optimize the mask prediction.

Unlike previous methods that directly decode masks using text features, our approach leverages the CMDecoder to deeply mine cross-modal associations between textual and visual cues, while simultaneously capturing semantic relationships among different targets. 
This enables the generation of more discriminative referring features, effectively addressing the challenges commonly encountered in remote sensing scenes.
By capturing these nuanced associations, our approach reduces mismatches between textual descriptions and visual targets, leading to more accurate alignment and segmentation.

% In contrast to region-based methods~\cite{wang2023caption,huang2024segment} that restrict outputs to the interior of the user box and overlook salient cues from surrounding context, our approach integrates both in-box evidence and global scene information to yield more accurate and context-aware results.

\subsection{Semantic and Relationship Consistency Learning} 
\label{sec:method_SRAlign}
\begin{figure}[t]
	\centering
	\includegraphics[width=1\linewidth]{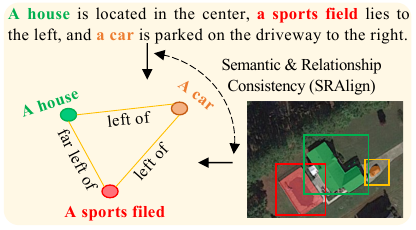}
	\caption{Semantic and Relationship Consistency Learning (SRAlign). SRAlign optimizes cross modal alignment by encouraging entity level semantic consistency and relationship level consistency, yielding captions and masks that are mutually faithful and coherent across modalities.}
    \label{fig:sralign}
\end{figure}

To improve both textual and visual decoding accuracy and consistency, we introduce the \textbf{Semantic and Relationship Consistency Learning (SRAlign)}, which enforces cross-modal alignment at the level of object semantics and inter-object relationships. As illustrated in ~\Cref{fig:sralign}, noun phrases in the generated caption should correspond to their segmentation masks in the image (e.g., the phrase ``a house'' aligns with the green mask covering the house), and the relationships described in text should mirror the spatial or semantic relationships among the segmented objects (e.g., ``a house is located between the sports field on the left and the car on the right'' should be reflected by the relative positions of the corresponding masks). Motivated by these observations, SRAlign comprises two complementary objectives: a Semantic Consistency Loss $\mathcal{L}_{\text{csc}}$, which promotes fine-grained semantic discrimination across modalities, and a Relationship Consistency Loss $\mathcal{L}_{\text{rec}}$ , which enforces structural coherence between textual relations and visual interactions. Together, these losses encourage precise, interpretable correspondences between caption tokens and segmentation masks.
%SRAlign jointly enforces semantic consistency and relationship consistency across modalities. By aligning semantic and relational information between visual and textual spaces, SRAlign leverages textual semantics to reduce ambiguity in visual pixel categories and uses visual cues to refine textual representations. This results in captions and masks that are mutually consistent and coherent across modalities.

\textbf{Cross-modal Semantic Consistency Loss.} 
Given a predicted mask-caption pair $(M, S)$, it is important to note that the relationship between masks and words is not strictly one-to-one. Therefore, the $i$-th mask embedding $\mathbf{m}_i$ is expected to be close to the matched word embedding sets ${\mathbf{s}_i }^+$, while maintaining a significant distance from other word embeddings $\mathbf{s}_j$. In other words, the mutual information between mask entities and their corresponding words should be maximized. However, directly optimizing mutual information is computationally infeasible due to its high cost and complexity. To address this, we introduce a cross-modal semantic consistency loss $\mathcal{L}_{\text{csc}}$ inspired by InfoNCE \cite{oord2018representation}, which provides a variational lower bound for mutual information optimization. This loss is formulated as follows:
\begin{equation}
\begin{split}
\mathcal{L}_{\text{csc}} = &\!- \sum_{i=1}^{N} \frac{1}{\vert \{ \mathbf{s}_i \}^+ \vert } \sum_{\mathbf{s}_i \in  \{ \mathbf{s}_i \}^+ } \!\!\!\!\!\log \frac{\exp(\mathbf{m}_i^\top  \mathbf{s}_i / \tau)}{\sum\limits_{j \neq i} \exp(\mathbf{m}_i^\top \mathbf{s}_j/\tau) } \\
&- \!\!\sum_{i=1}^{N} \frac{1}{\vert \{ \mathbf{m}_i \}^+ \vert} \sum_{\mathbf{m}_i \in \{ \mathbf{m}_i \}^+} \!\!\!\!\!\log \frac{\exp( \mathbf{s}_i^\top  \mathbf{m}_i / \tau)}{\sum\limits_{j \neq i} \exp(\mathbf{s}_i^\top  \mathbf{m}_j /\tau) } ,
\end{split}
\label{eq:7}
\end{equation}
where $\tau$ is a learnable parameter. Likewise, we impose that the $i$-th word embedding $\mathbf{s}_i$ is similar to the matched mask embedding sets $\{\mathbf{m}_i \}^+$ and dissimilar to the irrelevant $\mathbf{m}_j$.

\textbf{Relationship Consistency Loss.} 
The Relationship Consistency Loss is designed to align relational structures across modalities, ensuring consistency between object relationships in the visual space and their corresponding relationships in the textual embedding space. 
Given a predicted mask–caption pair $(M, S)$, we first extract the textual object representation $f^t_h$ from the last transformer layer of the LLM and compute the textual relationship matrix $\mathcal{R}^{t}_{\text{e}}$ for the objects referenced in (S) following Eqs.~\eqref{eq:mhca1}–\eqref{eq:4}.
Similarly, we obtain the visual object representation $f^v_h$ from a hidden layer of the SAM2 decoder and construct the corresponding visual relationship matrix $\mathcal{R}^{v}_{\text{e}}$ for the objects segmented in $M$ using the same procedure.
To encourage the model to learn consistent object relationships between the visual space and their counterparts in the textual embedding space, we minimize the Kullback–Leibler divergence between the two relationship distributions:

\begin{equation}
\begin{aligned}
\mathcal L_{\text{rec}}
&= \tfrac{1}{2}\Big[\mathrm{KL}(\mathcal R^t_e \| \mathcal R^v_e) + \mathrm{KL}(\mathcal R^v_e \| \mathcal R^t_e) \Big]\\
&= \frac{1}{2}\sum_{i=1}^{N}\sum_{j=1}^{N}\Big[
\mathcal{R}^{t}_{\text{e}}(i,j)\,\log\!\frac{\mathcal{R}^{t}_{\text{e}}(i,j)}{\mathcal{R}^{v}_{\text{e}}(i,j)}
\\
&\quad + \mathcal{R}^{v}_{\text{e}}(i,j)\,\log\!\frac{\mathcal{R}^{v}_{\text{e}}(i,j)}{\mathcal{R}^{t}_{\text{e}}(i,j)}
\Big].
\end{aligned}
\label{eq:8}
\end{equation}

The SRAlign loss $\mathcal{L}_{\text{SRAlign}}$ is formulated as:
\begin{equation}
\mathcal{L}_{\text{SRAlign}} = \mathcal{L}_{\text{rec}}+\mathcal{L}_{\text{csc}}. 
\label{eq:9}
\end{equation}
The minimization of $\mathcal{L}_{\text{SRAlign}}$ optimizes the alignment between corresponding mask-caption pairs by ensuring consistency in both semantic and relational information across modalities. This joint optimization enhances the model's ability to generate captions and masks that are not only accurate but also semantically and relationally aligned, thereby producing more coherent and consistent cross-modal outputs.

Finally, we train our model using either 
$\mathcal{L}_\text{Caption}$ or $\mathcal{L}_{\text{Mask}}$, combined with the proposed $\mathcal{L}_{\text{SRAlign}}$. The total loss is calculated as follows:
\begin{equation}
  \mathcal{L}_{\mathrm{total}} = \mathcal{L_{\text{Caption}}} + \mathcal{L_{\text{Mask}}} + {\lambda} \mathcal{L_{\text{SRAlign}}}.
\label{eq:total_loss}
\end{equation}
where ${\lambda}$ is a balancing weight.

\section{Experiments}  \label{experiments}

\subsection{Datasets and Metrics}
We conduct comprehensive experiments on both the remote sensing dataset GeoPixelD and the natural image dataset GranD. Specifically, GeoPixelD is employed to evaluate the controllable and relation-aware consistency of our method on high-resolution and object-dense remote sensing imagery, while GranD further demonstrates the generalizability of our approach to diverse natural image domains.

\textbf{GeoPixelD} is a multi-modal Grounded Conversation Generation (GCG) dataset comprising 53,816 grounded phrases aligned with 600,817 object masks, specifically designed for remote sensing image understanding. It provides hierarchically structured annotations that deliver rich semantic descriptions, seamlessly integrating comprehensive scene-level context with fine-grained, localized object-level details.

\textbf{GranD}  is explicitly designed for the GCG task in natural image domains and comprises approximately 214,000 image–grounded text pairs. Of these samples, 2,600 are reserved for validation and 5,000 for testing, with the remaining examples serving as training data to support large-scale model development and evaluation.

We evaluate our model using METEOR and CIDEr to assess caption quality, class-agnostic mask Average Precision (AP) to measure mask-to-phrase alignment, mask IoU to quantify mask quality, and mask recall to evaluate region-specific grounding performance.

\begin{table*}[t!]
    \centering
    \caption{To validate the effectiveness and generalization of our method, we provide a performance comparison with several advanced methods on the Grounded Conversation Generation Task using the remote-sensing dataset GeoPixelD and the natural-image dataset GranD. $\dagger$ denotes models that were fine-tuned on the training data. CLV-Net outperforms all other models across all metrics.}
    \renewcommand\arraystretch{1.0}
\centering
\resizebox{1\linewidth}{!}{
\begin{tabular}{l|ccccc|ccccc|c}
\toprule
\multirow{2}{*}{Model} &  \multicolumn{5}{c|}{GeoPixelD} & \multicolumn{5}{c|}{GranD} & \multirow{2}{*}{FPS} \\
                        & CIDEr & METEOR & AP50 & MIoU & RECALL & CIDEr & METEOR & AP50 & MIoU & RECALL \\ \midrule
LISA-7B$^\dagger$ $_{\text{(CVPR'24)}}$   & 14.6 & 22.3 & 8.5  & 42.7 & 29.0 & 33.9 & 13.0 & 25.2 & 62.0  & 36.3 & 0.61 \\
PixelLM-7B$^\dagger$ $_{\text{(CVPR'24)}}$  & 18.3 & 22.5  & 10.5   & 42.4   & 29.6 & 36.5 & 15.4 & 28.1 & 65.6   & 37.2 & 0.03 \\
GLaMM-7B$^\dagger$ $_{\text{(CVPR'24)}}$   & 15.7 & 23.0 & 12.5 & 46.4 & 32.8 & \textbf{47.2} & 16.2 & 30.8 & 66.3 & 41.8 & 0.09 \\
GeoPixel-7B $_{\text{(ICML'25)}}$ & 21.6 & 24.0 & 19.0 & 52.3 & 38.8 & 18.7 & 10.5 & 21.7 & 49.8 & 32.4 & 0.05 \\ 
CLV-Net-7B (ours) &  \textbf{24.5} & \textbf{26.2} & \textbf{21.8} & \textbf{55.3} & \textbf{41.9} & 47.0 & \textbf{16.8} & \textbf{33.4} & \textbf{69.2} & \textbf{43.6} & 0.08 \\  
\bottomrule
\end{tabular}
}

    \label{tab_1}
\end{table*}

\begin{figure*}[t!]
	\centering
\includegraphics[width=0.97\linewidth]{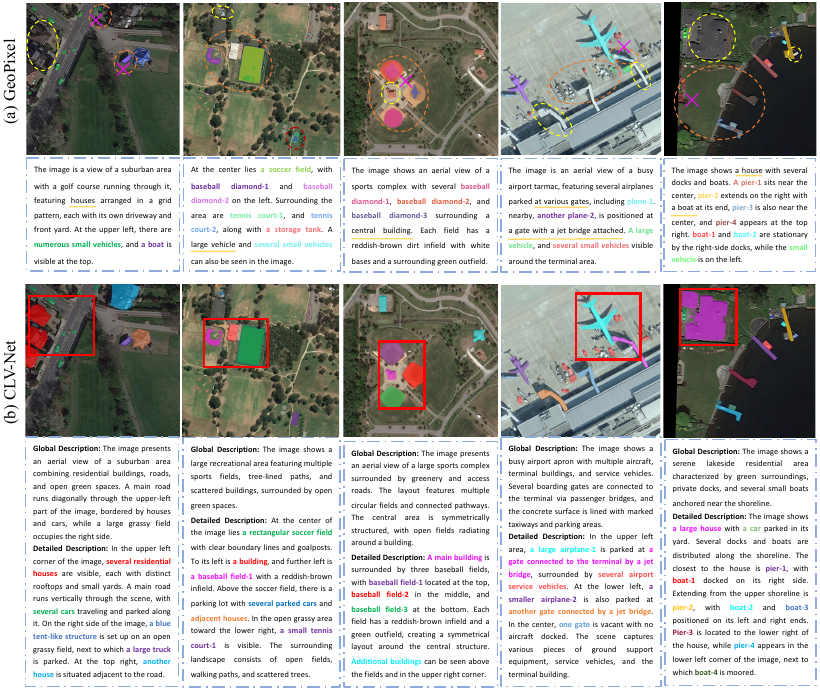}
	\caption{Quantitative results of our CLV-Net and GeoPixel on the GeoPixelD dataset, where the color indicates the matching relationship between masks and words. Underlined text denotes predicted words with missing masks, while crosses indicate mismatches between the generated text and the target.}
	\label{fig:results}
\end{figure*}

\subsection{Implementation Details} 
We adopt InternLM2.5-7B~\cite{zhang2024internlm} as the large language model for caption generation and employ LoRA~\cite{hu2022lora} for efficient fine-tuning.  
A fixed CLIP-ViT-L/14-336~\cite{radford2021learning} serves as the visual encoder, complemented by a grounded vision encoder initialized from SAM2~\cite{ravi2024sam} weights, while mask generation is performed by the SAM2 decoder. 
The trainable components of our architecture include the mask decoder ($\mathcal{D}$), LoRA parameters ($\alpha = 8$), and the CGFormer. 
In Eq.~\eqref{eq:total_loss}, we set $\lambda = 1$ to balance the optimization objectives. 
The learning rate is linearly warmed up to a maximum of $3 \times 10^{-4}$ over the first 150 training steps and then gradually decreased following a cosine decay schedule. The model is trained for 10 epochs with a batch size of $8$ using $4$ NVIDIA A6000 GPUs.

\begin{table}[t!]
    \centering
    \caption{Performance comparison of two CLV-Net variants, analyzing the impact of removing the prompt box and reducing the LLM model size.}
    \centering
\renewcommand\arraystretch{1.0}
\setlength\tabcolsep{2mm}
\resizebox{1\linewidth}{!}{
\begin{tabular}{lccccc}
\hline 
 Method    & CIDEr & METEOR & AP50 & MIoU & RECALL \\ \midrule
CLV-Net (7B) & \textbf{24.5}   & \textbf{26.2} & \textbf{21.8}  & \textbf{55.3} & \textbf{41.9}    \\
\hline
\rowcolor[gray]{.9} 
\multicolumn{6}{c}{{\textit{Without Prompt Box}}}   \\ 
CLV-Net (w/o Prompt) & 22.7   & 25.0 & 19.2  & 53.5 & 39.6     \\
\hline
\rowcolor[gray]{.9} 
\multicolumn{6}{c}{{\textit{Using Different LLMs Size}}}   \\
CLV-Net (1.8B) & 23.8   & 25.6 & 21.4  & 55.1 & 41.3    \\
\hline
\end{tabular}
}
    \label{tab_2}
\end{table}

\begin{table}[t!]
    \centering
    \caption{Performance analysis comparing the impact of different CMDecoder components on CLV-Net performance on the GeoPixelD dataset, including the effect of removing CGFormer (w/o CGFormer) and using different mask decoders.}
    \centering
\renewcommand\arraystretch{1.0}
\setlength\tabcolsep{2mm}
\resizebox{1\linewidth}{!}{
\begin{tabular}{lccccc}
\hline 
Method    & CIDEr & METEOR & AP50 & MIoU & RECALL \\ \midrule
CLV-Net (SAM2) & \textbf{24.5}   & \textbf{26.2} & \textbf{21.8}  & \textbf{55.3} & \textbf{41.9}    \\
\hline
\rowcolor[gray]{.9} 
\multicolumn{6}{c}{{\textit{Without CGFormer}}}   \\
CLV-Net (w/o CGFormer) & 23.2   & 25.6 & 20.5  & 53.0 & 39.4   \\ 
\hline
\rowcolor[gray]{.9} 
\multicolumn{6}{c}{{\textit{Using Different Mask Decoders}}}   \\
CLV-Net (SAM) & 24.4   & 26.0 & 21.2  & 54.8 & 41.5   \\ 
\hline
\end{tabular}
}
    \label{tab_3}
\end{table}

\subsection{Comparison with state-of-the-art methods}
We conduct comprehensive evaluations of our approach against leading state-of-the-art methods, including LISA~\cite{lai2024lisa}, PixelLM~\cite{ren2024pixellm}, GLaMM~\cite{rasheed2023glamm}, and GeoPixel~\cite{shabbir2025geopixel}, on both the remote sensing dataset GeoPixelD and the natural image dataset GranD. As summarized in \Cref{tab_1}, models are assessed using multiple metrics, including CIDEr, METEOR, AP50, mIoU, and recall. Our method consistently outperforms existing approaches, demonstrating robust performance and superior generalization across diverse diverse domains.

As shown in Table~\ref{tab_1}, methods designed specifically for natural image understanding exhibit clear limitations when applied to remote sensing data.
LISA$^\dagger$ performs worst on the remote sensing benchmarks due to its architectural limitations and its reliance on simple scene descriptions and coarse instance-level segmentation, which are poorly suited to remote sensing imagery. 
PixelLM$^\dagger$ achieves moderate improvements over LISA$^\dagger$ by incorporating multi-scale feature scaling and semantic token partitioning. 
GLaMM$^\dagger$ benefits from decoupled regional and global visual encodings and a unified multi-task pretraining strategy, which yields improvements over earlier baselines, but it does not explicitly address the distinct challenges posed by remote sensing data and therefore remains outperformed by GeoPixel and our method. 
GeoPixel, which is specifically tailored for remote sensing imagery, achieves substantial improvements compared with previous general-purpose models. Nevertheless, it still lags behind our approach. The performance gap is even larger on the natural image dataset, further highlighting the strong cross-domain generalization of our method.
Notably, we observe that on remote sensing datasets, natural-image-oriented models, even after fine-tuning, exhibit a significant performance gap compared with specialized remote sensing models. 
Conversely, on natural image datasets, GeoPixel underperforms compared with methods originally designed for natural scene understanding. 
In contrast, our approach not only achieves the best performance on the remote sensing benchmark but also reaches comparable or even superior results on natural image datasets. 
This advantage stems primarily from the ability of our method to capture fine-grained object semantics and to model relationships across modalities. Moreover, the model produces both globally coherent representations and finely detailed local outputs that align with user intent, thereby enhancing interpretable multimodal reasoning across domains.

\subsection{Qualitative Results} 
\Cref{fig:results} presents the visualization results of our method compared to GeoPixel~\cite{shabbir2025geopixel}. The yellow underline in the text correspond to the yellow dashed regions in the figure, indicating areas where the segmentation results fail to capture the targets described in the text. Cross marks in the image indicate incorrect matches between the segmented targets and the textual descriptions. 
In \Cref{fig:results} (a), GeoPixel generates a coarse, holistic description along with corresponding target masks, but it fails to capture most of the regional details and lacks a deep understanding of the scene. Within the yellow dashed regions, although the text specifies certain targets, the visual localization fails to identify them. In the red dashed regions, the text-related targets are located, but the segmentation quality is poor. In particular, the regions marked with crosses show incorrect matches between the target masks and the textual descriptions. These shortcomings significantly hinder the user's ability to accurately interpret the content of the image.
In contrast, as shown in \Cref{fig:results}(b), our method generates a global description, detailed local descriptions corresponding to regions of user interest, and high-quality target masks. The segmented masks exhibit precise boundaries, and the noun entities in the captions are accurately aligned with the corresponding objects in the image.
Overall, our approach leverages visual prompt-guided interaction to reduce user effort and effectively mitigate the challenges posed by the wide spatial coverage and high resolution of remote sensing imagery. 
Unlike region-based methods that constrain outputs to the interior of the user-defined box and ignore contextual cues from the surrounding area, our method integrates both in-box evidence and global scene information to produce more accurate and contextually aware results.

\subsection{Analysis of VPReasoner}
Our VPReasoner integrates local visual information with global features through a visual prompt box, guiding the large language model (LLM) to generate both a global description and fine-grained local details of interest to the user. To quantitatively evaluate the effectiveness of visual prompt guidance, we perform a comprehensive analysis that considers the impact of removing the visual prompt (w/o Prompt) as well as using LLMs of different sizes to examine how model scale affects output performance, as presented in \cref{tab_2}.

Without Prompt Box. Because our task emphasizes controllable outputs, the system must accept a user-provided visual prompt and then produce the user-desired results, thereby ensuring alignment with user intent. As shown in \cref{tab_2} (“w/o Prompt”), removing the prompt box produces a marked decline in both captioning and segmentation metrics. 
This result indicates that without explicit spatial guidance the model tends to default to global scene modeling and generates generic descriptions and masks that may include irrelevant background objects or omit the user-intended targets. These findings highlight the essential role of visual prompt guidance for producing precise, intent-aligned multimodal outputs.

Using Different LLMs.
We evaluated the impact of replacing the InternLM2.5-7B backbone with InternLM2.5-1.8B, and report the results in \cref{tab_2} (“Using Different LLMs”). Replacing 7B model with 1.8B variant yields a modest drop in captioning metrics while segmentation performance remains comparable. The performance change associated with model scale is substantially smaller than the degradation observed when the visual prompt is removed. These results indicate that the primary source of our performance gains arises from the newly proposed modules rather than from backbone model size.

\subsection{Analysis of CMDecoder}
Our CMDecoder employs an CGFormer to model inter-object dependencies and to bridge the visual and textual semantic gap, thereby producing enriched object representations. These enhanced representations are then passed to a mask decoder to generate the corresponding segmentation outputs.
To quantitatively evaluate the effectiveness of the CMDecoder, we perform a comprehensive ablation study that considers both the removal of CGFormer (w/o CGFormer) and the use of different mask decoders, as reported in \cref{tab_3}.

Effect of removing CGFormer.
The CGFormer in the CMDecoder explicitly captures relationships among objects and aligns visual features with textual semantics to improve segmentation fidelity and cross modal consistency. As shown in \cref{tab_3} (w/o CGFormer), removing the CGFormer causes declines in both captioning and segmentation metrics, with a larger drop observed in segmentation. This result indicates that richer, context-aware object representations substantially benefit visual decoding, and that optimizing these representations jointly also positively influences caption generation.

Using Different Mask Decoders. 
To assess the impact of the pretrained SAM2 model within our framework, we replace SAM2 with the SAM. As reported in \cref{tab_1} and \cref{tab_3}, the model equipped with SAM still outperforms the baselines. Although performance shows a modest decrease relative to the SAM2-based configuration, these results further confirm the effectiveness and generalizability of our framework.

\begin{figure*}[t!]
	\centering
\includegraphics[width=1\linewidth]{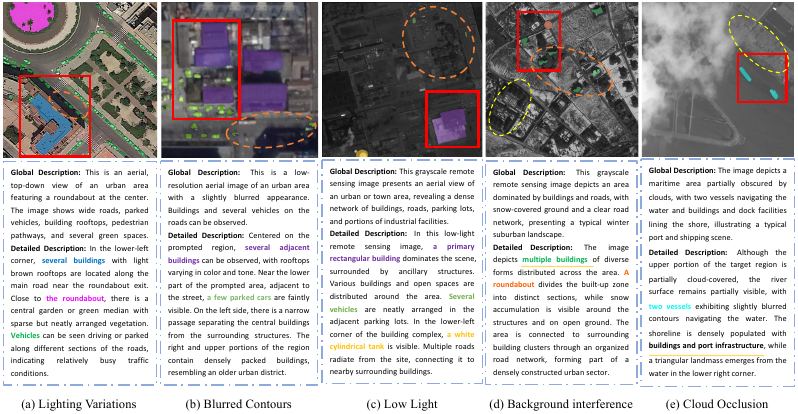}
	\caption{Visualization of several challenging and failure cases, where the colors indicates the matching relationship between masks and words. Red rectangles denote user-provided visual prompts; underlined text and yellow dashed circles denote predicted text with missing masks, while orange dashed circles indicate low-quality mask segmentation.}
	\label{fig:failure_cases}
\end{figure*}

\subsection{Analysis of SRAlign}
We conducted an ablation study to validate the effectiveness of the Semantic and Relationship Alignment Loss $\mathcal{L}_{\text{SRAlign}}$, which comprises two components, $\mathcal{L}_{\text{csc}}$ and $\mathcal{L}_{\text{rec}}$ (refer to Eq. \eqref{eq:7}-\eqref{eq:9}). The results are summarized in \cref{tab_4}. Both $\mathcal{L}_{\text{csc}}$ and $\mathcal{L}_{\text{rec}}$ yield substantial improvements across captioning and segmentation metrics.
Specifically, $\mathcal{L}_{\text{csc}}$ enforces object-level semantic alignment between predicted masks and its corresponding caption words, thereby establishing a robust cross-modal correspondence that enriches the semantic representation of objects and sharpens distinctions among different targets. 
Meanwhile, $\mathcal{L}_{\text{rec}}$ models inter-object relationships and captures valuable dependency cues among targets. By learning consistency of relationships across modalities, $\mathcal{L}_{\text{rec}}$ promotes mutual reinforcement between textual and visual decoding.
Joint optimization of these two components further enhances model performance, demonstrating the efficacy of the proposed $\mathcal{L}_{\text{SRAlign}}$. 
Additional evidence is provided in \cref{tab_5}, which examines the overall effectiveness of $\mathcal{L}_{\text{SRAlign}}$ under varying values of $\lambda$ in Eq. \eqref{eq:total_loss}. 
Optimal performance is achieved when $\lambda = 1$. Increasing $\lambda$ beyond this threshold reduces the contributions of both captioning and segmentation losses, leading to performance degradation. Conversely, reducing $\lambda$ weakens the alignment between captions and masks as well as the semantic relationships among targets, resulting in a corresponding decline in overall performance.

\begin{table}[t!]
    \centering
    \caption{Performance comparison of the impact of removing both or either of the losses, $\mathcal{L_{\text{rec}}}$ and $\mathcal{L_{\text{csc}}}$, in $\mathcal{L_{\text{SRAlign}}}$ on CLV-Net performance on the GeoPixelD dataset.}
    \centering
\renewcommand\arraystretch{1.0}
\resizebox{1\linewidth}{!}{
\begin{tabular}{cc|ccccc}
\hline 
$L_{csc}$    & $L_{rec}$    & CIDEr & METEOR & AP50 & MIoU & RECALL \\ \midrule
-            & -            & 23.7   & 25.3 & 19.6  & 53.8 & 40.2      \\
$\checkmark$ & -            & 24.2   & 25.6 & 21.5  & 55.1 & 41.3      \\
-            & $\checkmark$ & 24.0   & 25.8 & 21.3  & 54.9 & 41.0      \\ 
% \rowcolor[gray]{.9} 
$\checkmark$ & $\checkmark$ & \textbf{24.5}   & \textbf{26.2} & \textbf{21.8}  & \textbf{55.3} & \textbf{41.9}      \\ 
\hline
\end{tabular}
}

    \label{tab_4}
\end{table}

\begin{table}[t!]
    \centering
    \caption{Ablation study on the impact of different weightings of the $\mathcal{L_{\text{SRAlign}}}$ on the performance of CLV-Net.}
    \centering
\renewcommand\arraystretch{1.0}
\resizebox{0.95\linewidth}{!}{
\begin{tabular}{c|ccccc}
\hline
% \toprule
$\lambda$ & CIDEr & METEOR & AP50 & MIoU & RECALL  \\ \midrule
0       & 23.7   & 25.3 & 19.6  & 53.8 & 40.2       \\
% \rowcolor[gray]{.9} 
0.5       & 24.3   & 25.9 & 21.5  & 55.0 & 41.4     \\
1       & \textbf{24.5}   & \textbf{26.2} & \textbf{21.8}  & \textbf{55.3} & \textbf{41.9}       \\
5       & 23.9   & 25.4 & 21.2  & 54.6 & 41.1       \\
\hline
\end{tabular}
}
    \label{tab_5}
\end{table}

\subsection{Challenging Cases Visualization}
We visualize several challenging and failure cases in \cref{fig:failure_cases} for detailed analysis, including shadows (lighting variations), blurred contours, low light, background interference and cloud occlusion.
Our method produces imprecise segmentation contours in scenes where some targets are located in shadowed or lighting-variable areas. Similar failures occur in blurred images, as shown in \cref{fig:failure_cases}(a) and \cref{fig:failure_cases}(b).
In low-illumination scenes and those with strong background interference, targets exhibit low appearance contrast relative to their surroundings, which reduces discriminability and leads to incomplete or missed segmentations, as illustrated in \cref{fig:failure_cases}(c) and (d).
Furthermore, under snowy and foggy conditions our approach sometimes fails to segment target regions described in the text. We attribute these omissions to appearance changes induced by environmental interference that obscure salient features and impede reliable recognition, as shown in \cref{fig:failure_cases}(d) and \cref{fig:failure_cases}(e).

\subsection{Limitations and Future Perspectives}
Despite the promising performance of CLV-Net, several limitations remain that warrant further exploration. When the input data exhibit sensor deficiencies or environmental disturbances such as fog, snow, or low-light noise, the model often produces incomplete or imprecise segmentations. Its performance also degrades in complex scenarios characterized by heavy occlusion, illumination variation, or densely distributed targets, revealing limited robustness under diverse sensing and environmental conditions.
Future research will focus on enhancing the model’s adaptability through multimodal remote sensing data fusion, which can better capture complementary spatial and spectral information. In addition, developing self-supervised pretraining strategies may further improve generalization and reliability, enabling the model to perform more robustly across a wider range of real-world applications.

\subsection{Inference Speed and Model Parameters}
\label{sec:Complexity_Analysis}
In our CLV-Net framework, only the LoRA parameters and the CMDecoder are trainable, together accounting for merely 7.28\% of the total model parameters. This lightweight design substantially reduces both training and storage overhead while preserving strong representational capacity across modalities.
To further assess computational efficiency, we compare the inference speed of CLV-Net with several representative models, including LISA, PixelLM, GLaMM, and GeoPixel. As shown in \cref{tab_1}, CLV-Net achieves 0.08 FPS on an NVIDIA A6000 GPU, while LISA, PixelLM, GLaMM, and GeoPixel reach 0.61 FPS, 0.03 FPS, 0.09 FPS, and 0.05 FPS, respectively. 
Although the FPS of CLV-Net is lower than that of LISA, it remains competitive among grounded conversational generation (GCG) frameworks that perform multi-object reasoning.
Overall, our approach requires relatively few learnable parameters and supports efficient inference, achieving a favorable balance between model complexity, computational cost, and multimodal reasoning performance.

\section{Conclusion}  \label{conclusion}
In this work, we propose Cross-modal Context-aware Learning for Visual Prompt Guided Multimodal Image Understanding in Remote Sensing (CLV-Net), a framework designed to generate precise, user-tailored multimodal outputs by leveraging simple visual cues, such as bounding boxes, to guide attention toward regions of interest. CLV-Net integrates a Context-Aware Mask Decoder to model and incorporate inter-object semantic relationships, enhancing mask quality and target representations, and a Semantic and Relationship Consistency Learning (SRAlign) module that enforces cross-modal representation and relationship consistency while improving discrimination among visually similar targets through a combination of Cross-modal Semantic Consistency Loss and Relationship Consistency Loss.
Extensive experiments on both the remote sensing dataset GeoPixelD and the natural image dataset GranD demonstrate that CLV-Net consistently outperforms existing state-of-the-art methods. The framework effectively captures user intent, resolves fine-grained visual ambiguities, and produces high-quality, aligned caption and mask outputs across diverse visual domains.

{
\bibliographystyle{IEEEtran}
\bibliography{main}
}

\end{document}